\documentclass[times, review, 10pt]{elsarticle}

\usepackage{setspace} % 引入 setspace 宏包
\usepackage{lineno,hyperref}
\usepackage{amssymb}
\usepackage{amsmath}
\usepackage{amsfonts}
\usepackage{graphicx}
\usepackage{array}
\usepackage{booktabs}
\usepackage{float}
\usepackage{multirow}
\usepackage{color}
\usepackage[ruled]{algorithm2e}
\usepackage{subfigure}
\usepackage{bbding}
\usepackage{hyperref}
\usepackage{diagbox}
\usepackage{hyperref}
\usepackage{amsfonts,amssymb} 
\usepackage{soul, color, xcolor}
 \usepackage[nodots]{numcompress}
\hypersetup{hidelinks,
	colorlinks=true,
	allcolors=black,
	pdfstartview=Fit,
	breaklinks=true}
\modulolinenumbers[5]
\biboptions{sort&compress}
\journal{Journal of \LaTeX\ Templates}

\begin{document}

\begin{frontmatter}

\title{Global and Local Mamba Network for Multi-Modality Medical Image Super-Resolution}

\author[mymainaddress,mysecondaryaddress,myforthaddress]{Zexin Ji}
\author[mymainaddress,mysecondaryaddress]{Beiji Zou}
\author[mymainaddress,mysecondaryaddress]{Xiaoyan Kui\corref{mycorrespondingauthor}}\cortext[mycorrespondingauthor]{Corresponding author}
\ead{xykui@csu.edu.cn}
\author[mythirdaddress,myforthaddress]{Sbastien Thureau}
\author[mythirdaddress]{Su Ruan}

\address[mymainaddress]{School of Computer Science and Engineering, Central South University, Changsha, 410083, China}
\address[mysecondaryaddress]{Hunan Engineering Research Center of Machine Vision and Intelligent Medicine, Central South University, Changsha, 410083, China}
\address[mythirdaddress]{University of Rouen-Normandy, LITIS - QuantIF UR 4108, F-76000, Rouen, France}
\address[myforthaddress]{Department of Nuclear Medicine, Henri Becquerel Cancer Center, Rouen, France}

\begin{abstract}

Convolutional neural networks and Transformer have made significant progresses in multi-modality medical image super-resolution. However, these methods either have a fixed receptive field for local learning or significant computational burdens for global learning, limiting the super-resolution performance. To solve this problem, State Space Models, notably Mamba, is introduced to efficiently model long-range dependencies in images with linear computational complexity. Relying on the Mamba and the fact that low-resolution images rely on global information to compensate for missing details, while high-resolution reference images need to provide more local details for accurate super-resolution, we propose a global and local Mamba network (GLMamba) for multi-modality medical image super-resolution. To be specific, our GLMamba is a two-branch network equipped with a global Mamba branch and a local Mamba branch. The global Mamba branch captures long-range relationships in low-resolution inputs, and the local Mamba branch focuses more on short-range details in high-resolution reference images. We also use the deform block to adaptively extract features of both branches to enhance the representation ability. A modulator is designed to further enhance deformable features in both global and local Mamba blocks. To fully integrate the reference image for low-resolution image super-resolution, we further develop a multi-modality feature fusion block to adaptively fuse features by considering similarities, differences, and complementary aspects between modalities. In addition, a contrastive edge loss (CELoss) is developed for sufficient enhancement of edge textures and contrast in medical images. Quantitative and qualitative experimental results show that our GLMamba achieves superior super-resolution performance on BraTS2021and IXI datasets. We also validate the effectiveness of our approach on the downstream tumor segmentation task.

\end{abstract}

\begin{keyword}
Multi-Modality, Magnetic Resonance Imaging, Super-Resolution, Mamba, Deformable, Tumor Segmentation
\end{keyword}

\end{frontmatter}

%\linenumbers

\section{Introduction}

Medical imaging techniques like magnetic resonance imaging (MRI) are essential tools for clinical diagnosis. High-resolution medical images can provide radiologists with clearer anatomical and pathological information, thereby enabling more accurate diagnoses. However, obtaining high-resolution medical images relies on longer scanning time and costly medical equipment. To reduce costs, medical image super-resolution (SR) has attracted significant attention in the field of medical imaging. The objective of this technique is to generate high-resolution (HR) images from low-resolution (LR) versions in a cost-effective way~\cite{DBLP:journals/mia/XiaRGNPF21}. The current medical image super-resolution can be broadly divided into two categories: single-modality SR approaches and multi-modality SR approaches. Single-modality SR approaches focus on enhancing the resolution of a single type of medical imaging but overlook the auxiliary information from multi-modality images. 
Multi-modality SR approaches can integrate multi-types of information coming from multiple imaging modalities to achieve more accurate image super-resolution. In clinical practice, different imaging modalities for the same subject can provide distinct types of diagnostic information.
For example, T1-weighted (T1) images primarily highlight anatomical details, while T2-weighted (T2) images emphasize pathology. During the acquisition process, T2 images typically have a longer acquisition time than T1. Therefore, we can use the modality with a shorter acquisition time as a reference image to provide auxiliary information for the modality with a longer acquisition time. This can reduce acquisition time while enhancing image resolution using reference information. Specifically, T1 images can be used to improve the resolution of T2 images.

Benefiting from the strong capability of convolutional neural networks (CNNs) to learn the nonlinear mapping between low-resolution and high-resolution images, substantial advancements~\cite{10,8,11,DBLP:journals/cbm/ZengZCYZC18} have been achieved for single- and multi-modality medical image super-resolution. However, such CNN-based methods only leverage local information from a limited receptive field, while neglecting the global tissue structure in medical images.
The vision Transformer (ViT) has gained prominence in medical image super-resolution task~\cite{DBLP:journals/bspc/ZouJZDZK23,forigua2022superformer,fang2022cross,DBLP:conf/iccv/LiZSLZCLLX23}, showcasing its ability to capture global contexts information. For single-modality MR image super-resolution,
Forigua \emph{et al.}~\cite{forigua2022superformer} introduced SuperFormer to employ the Swin Transformer~\cite{liu2021swin}.
Huang \emph{et al.}~\cite{DBLP:journals/cmig/HuangLCHJW24} proposed a CNN-ViT framework to employ convolutions with adaptive kernels for local feature extraction, while global context is captured through non-local mechanisms and vision transformers. Hua \emph{et al.}~\cite{DBLP:journals/bspc/HuaDMY24} developed MRI super-resolution network built on three modules: convolutional layers and cross-iterative structures enrich local features, and a sparse attention graph neural network (GNN) connects distant pixels to strengthen global understanding. Both two approaches follow a shared design philosophy, using convolutional operations for local representation and Transformer or GNN modules to model global dependencies in single-modality super-resolution tasks. For multi-modality MR image super-resolution,
Li \emph{et al.} proposed McASSR~\cite{DBLP:conf/iccv/LiZSLZCLLX23} using cross-attention transformers and reference-aware implicit attention for upsampling and information fusion. 
Feng \emph{et al.}~\cite{feng2024exploring} designed SANet with separable attention and multi-stage integration, leveraging auxiliary contrast images to focus on both high- and low-intensity regions and enhance structural clarity and detail accuracy. 
Despite these advances, existing methods still have the following limitations: (1) Trade-off between representation capability and computational cost.Convolutional neural networks (CNNs), while efficient in capturing local features, are limited in their ability to model long-range dependencies. Transformer-based architectures address this limitation with strong global modeling capabilities.The calculation of self-attention demands space and time resources that grow quadratically with the increase in the number of tokens. (2) Underutilization of Intra-modality information. Existing multi-modal methods typically treat low-resolution images and high-resolution reference images equally, without fully considering their distinct roles in super-resolution process. Low-resolution medical images, while lacking fine anatomical details, still retain valuable global structural information that remains underexploited by current approaches. In contrast, high-resolution reference images provide richer local textures and detailed anatomical cues, yet their locally correlated nature is often overlooked. (3) Inadequate exploitation of inter-modality dependencies. Existing approaches primarily focus on the complementary relationship between low-resolution and high-resolution reference modality, but pay insufficient attention to the inherent similarities and differences across modalities. Fully capturing these inter-modality relationships is essential for accurate feature fusion and to minimize information loss during the super-resolution process.

Recently, State Space Models (SSMs)~\cite{DBLP:conf/nips/GuJGSDRR21} have gained significant attention for their strong performance in sequence modeling. A new SSM architecture called Mamba~\cite{DBLP:journals/corr/abs-2312-00752} was introduced, incorporating a selective scan state space model with data-driven mechanisms and efficient hardware design. Unlike CNNs, which rely on deep stacks to expand the receptive field, or Transformers, which require heuristic window partitioning with substantial parameter overhead to model long-range dependencies, the selective state space mechanism (S6) in Mamba enables efficient dynamic integration of global features with linear complexity. MRI super-resolution faces challenges in recovering complex anatomical structures such as cortical folding and tissue boundaries due to their low contrast and widespread distribution. Mamba leverages state-space modeling to capture long-range dependencies and dynamically adjust information flow based on spatial context. Its linear complexity enables fast inference, meeting the demands of clinical high-resolution imaging. Therefore, it is desirable to develop a Mamba-based SR approach to explore the feature interaction in medical images by regrouping the feature learning scope.

Based on the above analysis, we propose a global and local Mamba network, named GLMamba for multi-modality medical image super-resolution (Fig.~\ref{fig:Framework}), which aims to thoroughly explore feature learning both within and across modalities. 
Our GLMamba is a two-branch network, including a global Mamba branch for low-resolution image and a local Mamba branch for high-resolution reference image. Low-resolution images require global information to compensate for missing details, while high-resolution reference images need detailed local features to assist the accuracy of super-resolution generation. Therefore, the global Mamba branch is tasked with capturing global relationships among image patches for low-resolution input images. The local Mamba branch is tailored for exploring short-range local patch clues for high-resolution reference by dividing the whole image into four quadrants. The deform block is further employed to enhance the representation ability of the model by extracting adaptive features from pixel regions for two branches. The modulator serves to further refine and enhance pixel-level deformable features within patch-wise global LR features and local Ref features, thereby improving the perception of fine-grained details before feature integration.
Moreover, the local information in the reference image can provide a guide to low-resolution image. In light of this, we design a multi-modality feature fusion block to adaptively fuse features by considering similarities, differences, and complementary aspects between the modalities. To better capture high-frequency details in super-resolved images, we introduce a contrastive edge loss (CELoss) that specifically targets the enhancement of edge textures and contrast in medical images. The quantitative and qualitative super-resolution results demonstrate the effectiveness of our approach. Additionally, we also validate our approach on the downstream tumor segmentation task.

Specific contributions of our work are summarized as follows: 1) We develop a global and local Mamba network for multi-modality medical image super-resolution, to the best of our knowledge, this is the first method that explores the potential of both global and local Mamba information for multi-modality medical image super-resolution. 2) We introduce a Deform block to capture fine-grained pixel-level structures from low-resolution and reference images. A modulator is employed to further refine these features within patch-wise global and local contexts for improved spatial detail perception before fusion. 3) We design a multi-modality feature fusion block to adaptively fuse the high-resolution reference information from similarities, differences, and complementary aspects. 4) We propose a contrastive edge loss to improve the preservation of edge textures and contrast details in super-resolved images.

\section{Related Works}\label{Related Works}

\subsection{Single-Modality Super-Resolution}

Medical image super-resolution aims to generate high-resolution medical images from corresponding low-resolution version. A range of methods~\cite{DBLP:journals/cmig/PhamTMBFPR19,DBLP:journals/cmig/MahapatraBG19,DBLP:journals/pr/DingZXW24,DBLP:journals/pr/SuLXFL25} leverage CNNs to learn the powerful feature extraction capabilities to capture complex patterns and details, enabling more accurate and visually appealing super-resolution results compared to traditional techniques.
For example, Phama \emph{et al.}~\cite{DBLP:journals/cmig/PhamTMBFPR19} proposed a residual-based deep 3D CNN architecture for brain MRI super-resolution. Mahapatra \emph{et al.}~\cite{DBLP:journals/cmig/MahapatraBG19} introduced a progressive GAN (P-GAN) for medical image super-resolution, which generates high-resolution images at each stage using a triplet loss.
Although these CNN-based approaches demonstrate impressive performance, the reliance on local receptive fields inherently biases them towards local features.
After that, Transformers have gained attention in the field of medical image super-resolution due to their ability to effectively capture long-range dependencies of images. Transformers utilize self-attention mechanisms to model relationships across the entire image, enabling the reconstruction of finer details and more accurate high-resolution outputs.
Forigua \emph{et al.}~\cite{forigua2022superformer} proposed a volumetric transformer-based approach for MRI super-resolution. Liang \emph{et al.} presented SwinIR~\cite{DBLP:conf/iccvw/LiangCSZGT21} based on the Swin Transformer~\cite{DBLP:conf/iccv/LiuL00W0LG21} to feature shallow and deep feature extraction and high-quality image reconstruction.
Transformer-based methods for medical image super-resolution are promising but face challenges like high computational demands due to self-attention mechanisms and the need for large datasets. 
Recently, Mamba, characterized by linear computational complexity for long-range dependency calculations, has been successfully introduced into the study of medical image super-resolution.
Ji \emph{et al.}~\cite{DBLP:journals/prl/JiZKLVR25} designed the self-prior guided MambaUnet network to 
leverage the self-similarity information of medical images as a prior to integrate multi-scale global information for MR image super-resolution.
However, the aforementioned methods generally prioritize information from a single modality when restoring high-resolution images, thereby neglecting the complementary information available across different modalities in medical imaging.

\subsection{Multi-Modality Super-Resolution}
In clinical settings, different imaging modalities used on the same patient are necessary to provide complementary information to aid diagnosis. Therefore, multi-modality super-resolution~\cite{DBLP:conf/miccai/FengFYX21,DBLP:conf/cvpr/LiLTDWX022,DBLP:conf/bibm/JiKLZLDZ23,feng2024exploring,DBLP:journals/eaai/YanWCSZZLJZ24} can further enhance low-resolution image resolution by incorporating effective information from high-resolution reference modalities. 
Feng \emph{et al.}~\cite{DBLP:conf/miccai/FengFYX21} presented a multi-stage integration network (MINet) for multi-contrast MRI super-resolution, designed to explicitly capture and utilize the dependencies between different contrast images at various stages to enhance the super-resolution process.
Li \emph{et al.}~\cite{DBLP:conf/cvpr/LiLTDWX022} proposed Transformer-based multi-scale contextual matching and aggregation methods for multi-modality super-resolution.
Ji \emph{et al.}~\cite{DBLP:conf/bibm/JiKLZLDZ23} designed a wavelet-aware transformer network to dynamically integrate the complementary details of the multi-modality images within the wavelet domain and further enhance in the image domain.
Feng \emph{et al.}~\cite{feng2024exploring} introduced a separable attention network, dubbed SANet, which consists of high-intensity priority attention and low-intensity separation attention mechanisms to effectively fuse multi-modality information. While these methods have delivered impressive results, they are limited to high computational demands and cannot be processed in a lightweight manner, making them less ideal for clinical needs.

\subsection{Theoretical knowledge of Mamba}

Recently, state space models (SSMs) integrated into deep learning have garnered significant attention. The Structured State Space Sequence model (S4)~\cite{DBLP:conf/iclr/GuGR22} represents a major advancement in deep state space models, providing a static representation that is independent of the content. Building upon S4, a new SSM called Mamba has been proposed. Mamba integrates the Selective Scan State Sequence model (S6), which incorporates a selective mechanism and is designed to optimize hardware efficiency. Compared to Transformers, Mamba offers the advantage of modeling long-range dependencies while scaling linearly with sequence length. 
Mamba transforms a one-dimensional sequence $x(t) \in \mathbb{R}$ into the output $y(t) \in \mathbb{R}$ by utilizing a hidden state $h(t) \in \mathbb{R}^{\mathbb{N}}$, typically realized through the following linear ordinary differential equations (ODEs).
\begin{equation}\label{eq1}
    h^{\prime}(t)=\mathbf{A} h(t)+\mathbf{B} x(t), y(t)=\mathbf{C} h(t),
\end{equation}
where $\mathbf{A} \in \mathbb{R}^{\mathrm{N} \times \mathrm{N}}$ are the state matrix. $\mathbf{B} \in \mathbb{R}^{\mathbb{N} \times 1}$ and $ \mathbf{C} \in \mathbb{R}^{1 \times \mathbb{N}}$ are the parameters of projection.

The Zero-Order Hold (ZOH) method is a commonly employed technique for converting ODEs into discrete functions, making it particularly suitable for deep learning applications. By introducing a timescale parameter $\Delta$, this approach facilitates the conversion of continuous-time system matrices, represented by $\mathbf{A}$ and $\mathbf{B}$, into their discrete equivalents $\overline{\mathbf{A}}$ and $\overline{\mathbf{B}}$. This conversion preserves the original system's dynamics and ensures compatibility with discrete computational frameworks in deep learning. Mamba improves traditional SSMs by deriving parameters $\overline{\mathbf{B}}$, $\mathbf{C}$, and $\Delta$ directly from the input data, enabling the model to adapt to each input. The discretization process is carried out as follows:
\begin{equation}\label{eq2}
\overline{\mathbf{A}}=\exp (\Delta \mathbf{A}), \overline{\mathbf{B}}=(\Delta \mathbf{A})^{-1}(\exp (\Delta \mathbf{A})-\mathbf{I}) \cdot \Delta \mathbf{B} .
\end{equation}

Following discretization, Equation \ref{eq1} is converted into a format suitable for discrete-time processing, as illustrated below:
\begin{equation}\label{eq3}
h_t=\overline{\mathbf{A}} h_{t-1}+\overline{\mathbf{B}} x_t, y_t=\mathbf{C} h_t .
\end{equation}

Finally, the model produces the output through a global convolution as described below:
\begin{equation}\label{eq4}
\overline{\mathbf{K}}=\left(\mathbf{C} \overline{\mathbf{B}}, \mathbf{C} \overline{\mathbf{AB}}, \ldots, \mathbf{C} \overline{\mathbf{A}}^{{M}-1} \overline{\mathbf{B}}\right), \mathbf{y}=\mathbf{x} * \overline{\mathbf{K}},
\end{equation}
where $M$ is the length of the input sequence $\mathbf{x}$. $\overline{\mathbf{K}} \in \mathbb{R}^{\mathrm{M}}$ are structured convolutional kernel.

\subsection{Applications of Mamba}

The breakthrough of Mamba has attracted widespread attention in the field of computer vision and has been applied to various visual tasks, positioning vision Mamba as a novel alternative to CNNs and Transformers.
For example, 
Zhu \emph{et al.}~\cite{zhu2024vision} introduced Vision Mamba (Vim), integrating bidirectional SSM for modeling global visual context.
Ma \emph{et al.}~\cite{DBLP:journals/kbs/MaW24} proposed a purely Mamba-based U-shaped encoder-decoder network named Semi-Mamba-UNet for semi-supervised medical image segmentation.
Li \emph{et al.}~\cite{DBLP:journals/tgrs/LiLZWD24} developed the MambaHSI for
hyperspectral image classification, which contains spatial Mamba block for pixel-level long-range interactions and spectral Mamba block for spectral group feature extraction.

For the MRI super-resolution task, unique challenges arise compared to natural image restoration. These stem from the need to accurately reconstruct critical anatomical structures, such as cortical folding and tissue boundaries, which are widely distributed and often exhibit low contrast or weak signals. The state-space modeling paradigm of Mamba effectively captures long-range spatial dependencies by continuously maintaining and updating hidden states over extended contexts, thereby facilitating the accurate recovery of these complex anatomical details. Importantly, Mamba possesses dynamic modeling capability through its state-update mechanism that automatically adjusts information propagation and update strategies based on spatial position. This enables Mamba to flexibly modulate its memory length and focus of attention, rapidly aggregating features in simpler regions while dedicating more modeling capacity to complex anatomical boundaries. Finally, Mamba maintains computational efficiency with linear complexity and supports clinical workflows that require high-resolution outputs and fast inference. 
Ji \emph{et al.}~\cite{DBLP:conf/miccai/JiZKVR24} designed Deform-Mamba architecture for MR image super-resolution. They also proposed a self-guided Mamba network with edge-aware constraint~\cite{DBLP:journals/prl/JiZKLVR25} for medical image super-resolution. 
Our approach extends from single-contrast to multi-modality medical image super-resolution, proposing a global and local Mamba with inter-modal feature fusion to capture both modality-specific and cross-modal interactions. Compared to the U-Net-based design in prior work, we adopt a fully Mamba-based lightweight framework.

\section{Method}\label{Method}

\begin{figure}[t]
    \centering
    \includegraphics[width=1\linewidth]{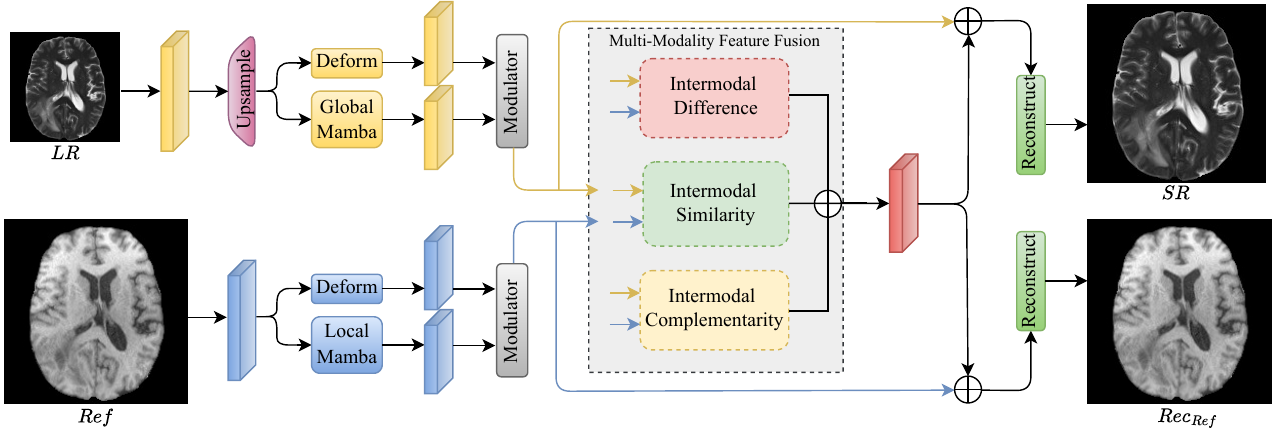}
    \caption{The overall pipeline of our approach consists of two branches, i.e., global Mamba branch and local Mamba branch. 
    The Deform block can adaptively extract deformable features from both branches. A modulator is introduced to further refine the deformable features within both the global and local Mamba blocks.
    The multi-modality feature fusion block is designed to fuse two branches. Super-resolved image ($SR$) and reconstructed reference image ($Rec_{Ref}$) are finally obtained through the reconstruct convolution layer.
    }
    \label{fig:Framework}
\end{figure}

\begin{figure}[t]
    \centering
    \includegraphics[width=0.9\linewidth]{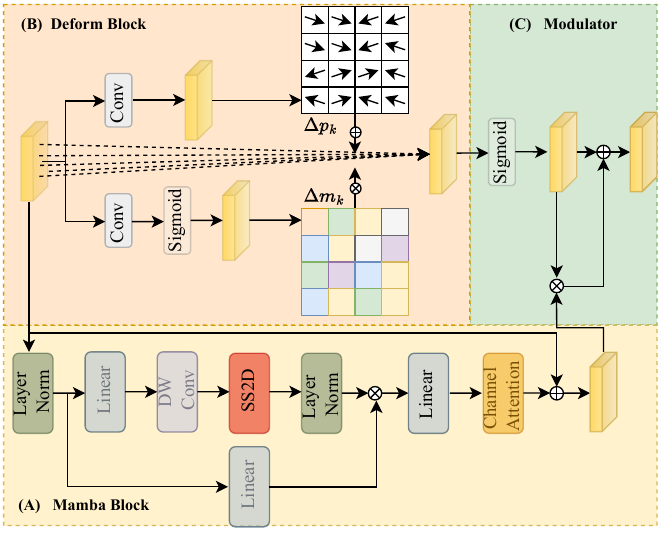}
    \caption{Architecture of the Mamba block (A), deform block (B) and the modulator (C).}
    \label{fig:Mamba}
\end{figure}

\subsection{Overall Architecture}

Fig.~\ref{fig:Framework} illustrates the pipeline of our proposed architecture.
$LR$ and $Ref$ represent the input low-resolution image and the high-resolution reference image, respectively.
For the $LR$, the upsampling layer is first applied to match the size of the reference image, which is then fed into the global Mamba block and deform block. For the $Ref$, we input it into the local Mamba block and deform block.
Specifically, we utilize the deform block to dynamically adjust the sampling strategy based on input feature content, facilitating the extraction of features from pixel region~\cite{DBLP:conf/iccv/DaiQXLZHW17}. However, its ability to capture global information is limited. To mitigate this, we also introduce a Mamba block.
Low-resolution images inherently lack detailed information, making it essential for a global Mamba to integrate the overall semantic and structural information, thereby facilitating a deeper understanding of the global features. In contrast, high-resolution reference images are rich in details and textures. Thus, applying a local Mamba allows for the precise capture of fine-grained details and short-range dependencies, enabling meticulous processing of intricate structures and textures from patch regions.
We further design a modulator to enhance the interaction between deformable features and the Mamba features for both branches, respectively.
Inspired by~\cite{DBLP:conf/cvpr/ZhangGJZH23}, we propose to fully fuse the feature of multi-modality by considering the similarities, differences, and complementary aspects.
These fused features are further aggregated by a reconstruct convolution layer to generate the super-resolved image $SR$ and reconstructed reference image $Rec_{Ref}$, respectively. 
Moreover, we also design a contrastive edge loss (CELoss) to constrain the edge generation and contrast of the super-resolved image.
Details will be given below.

\subsection{Global and Local Mamba}

The Mamba block operates at the patch level, leveraging state-space models to enable sequential information interaction and long-range dependency modeling. Unlike Deform block, which is still constrained by an adaptive convolutional window, Mamba enables effective information propagation across broader spatial extents. This allows it to capture global contextual cues and mid- to long-range structural relationships.
The Deform block focuses primarily on pixel-level flexible sampling and spatially adaptive feature aggregation. By learning dynamic offsets at each position, it enables fine-grained spatial modeling of the input image, effectively handling geometric deformations and local contextual variations. This makes it especially adept at capturing irregular or non-uniform local and semi-local patterns.

While both Deform and Mamba block address spatial dependencies, they differ in modeling scale, operational granularity, and functional focus, forming a complementary relationship:
(1) The combination of Global Mamba and Deformable achieves an effective integration of global and local information for low-resolution input image. Global Mamba constructs the macro-level structural framework and captures broad spatial relationships, while Deformable compensates for the lack of local detail, enhancing the representation of edges and microstructures. Together, they allow the model to balance overall layout understanding with fine-grained precision, significantly improving the reconstruction quality of low-resolution inputs.
(2) The combination of Local Mamba and Deformable for high-resolution reference image, though both emphasize local perception, offers distinct yet complementary strengths. Local Mamba enhances local contextual dependencies through sequential modeling, improving the coherence and structural consistency of features in the reference image. Deformable focuses on pixel-level flexible alignment and adaptation to local micro-level variations. In multi-contrast MRI super-resolution, where the reference and input images have subtle differences in contrast and structure, this combination enables precise cross-modal complementary feature enhancement, fully exploiting the detailed information in the reference image and further boosting super-resolution quality.

\begin{figure}[t]
    \centering
    \includegraphics[width=1\linewidth]{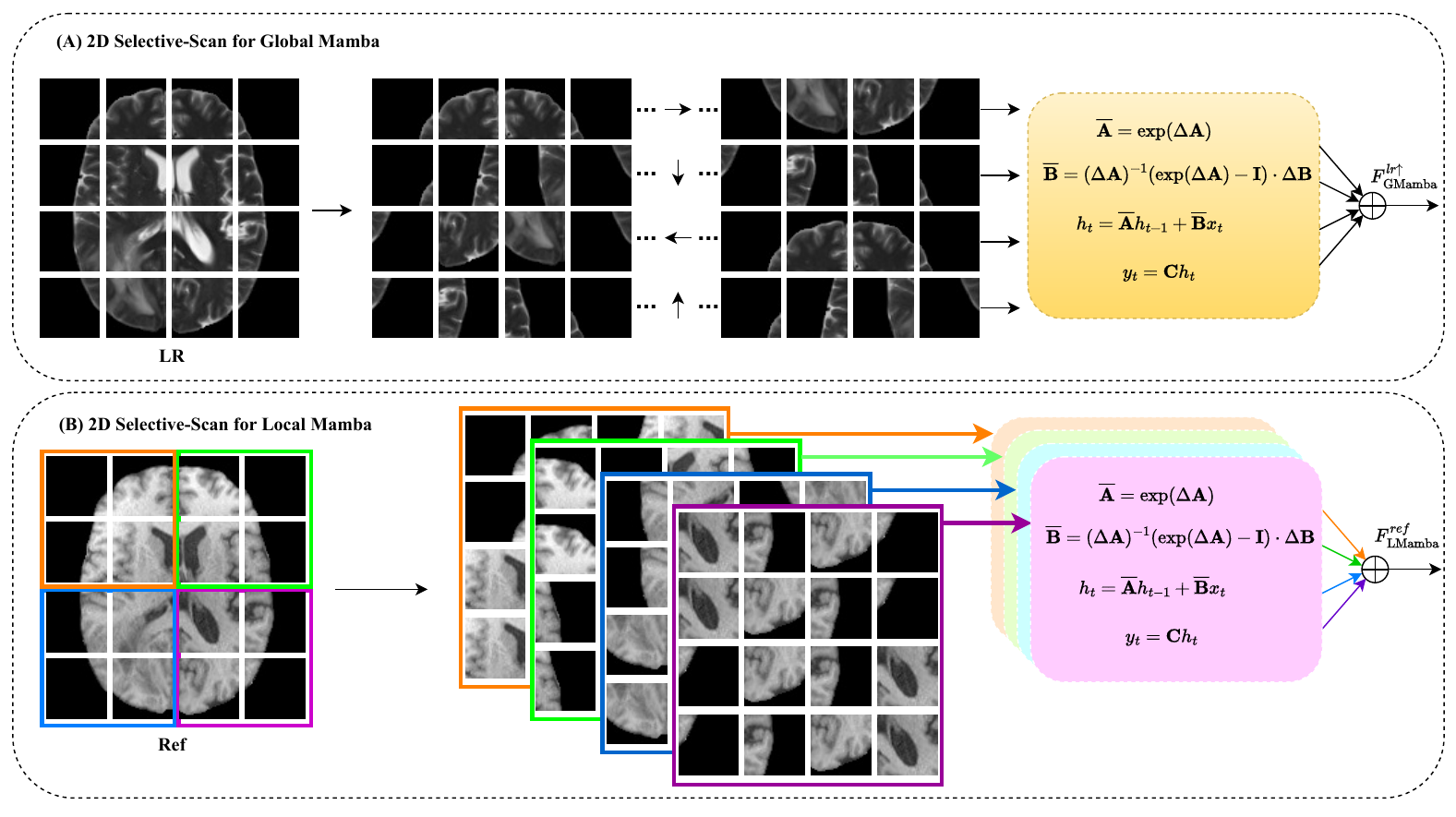}
    \caption{Architecture of the SS2D for global Mamba (A) and local Mamba (B). $F_{\text {GMamba}}^{lr\uparrow}$ and $F_{\text {LMamba}}^{ref}$ are the output features of SS2D for global and local Mamba, respectively.}
    \label{fig:SS2D}
\end{figure}

\textit{1) Global and Local Mamba Block:}
The framework of the Mamba block is illustrated in Fig.~\ref{fig:Mamba}(A). The input feature map is divided into two paths after undergoing layer normalization. The first path processes the input through a linear layer and an activation function. The second path sequentially passes through a linear layer, depthwise separable convolution, and an activation function before entering the 2D Selective Scan (SS2D) and layer normalization. Finally, these two paths are integrated through a multiplication operation. The channel attention mechanism used at the end further captures the inter-channel relationships.
The core of Mamba is the 2D Selective Scan (SS2D).
Fig.~\ref{fig:SS2D}(A) illustrates the detailed implementation of SS2D for global Mamba block.
Given that low-resolution images require global information to compensate for missing details, we employ a global Mamba for feature learning.
High-resolution reference images contain rich detailed information, and the local features captured by the local Mamba can more effectively capture the correlations between neighboring patches. Therefore, we divide the image into four quadrants to independently learn the detailed local correlations between patches within each quadrant (Fig.~\ref{fig:SS2D}(B)).
Similar to the Transformer, Mamba also divides the input image into patches. However, the image patches lack direct inferential relationships. Therefore, we use scan expanding operation to unfold the image into sequences in four different directions: from left to right, from top to bottom, from right to left, and from bottom to top. The purpose is to fully exploit the correlations between patches in different directions. The four sequences are further processed by equations shown in Fig.~\ref{fig:SS2D} (B) to output enhanced features. Finally, the scan merging operation combines the features from the different sequences and restores them to the original size.

\textit{2) Deform Block:}
Fig.~\ref{fig:Mamba}(B) illustrates the network structure of the deform block. Unlike traditional convolution, which uses a fixed sampling pattern, the deform block dynamically learns the convolution offsets based on the image content. This allows it to adaptively extract features more effectively. The operation can be described as follows:
\begin{equation}
Y(p)=\sum_{k=1}^K w_k \cdot X\left(p+p_k+\Delta p_k\right) \cdot \Delta m_k,
\end{equation}
where $k$ is the index. $K$ is the number of convolution kernels. $w_k$ represents the $k$-th kernel weight, and $X$ and $Y$ denote the input and output feature map, respectively. The terms $p$, $p_k$, $\triangle p_k$ and $\triangle m_k$ represent the sampling position, the $k$-th predefined offset, dynamic offset, and scalar for the $k$-th location.
In particular, $\Delta p_k \in \mathbb{R}^{H \times W \times 2 C}$ and $\Delta m_k \in \mathbb{R}^{H \times W \times C}$ are learned by the convoluional layer. 
Specifically,
$\Delta p_k=\operatorname{Conv}\left(\mathbf{X}\right)$, and $\Delta m_k=\sigma\left(\operatorname{Conv}\left(\mathbf{X}\right)\right)$, where $\sigma$ is the sigmoid function.
The learned $\triangle p_k$ is then added to $p_k$, allowing the sampling position to adapt dynamically based on the content of the image. The dashed lines in Fig.~\ref{fig:Mamba}(B) represent the sampling paths of the convolution kernels after dynamic adjustment through offsets. 
Furthermore, $\triangle m_k$ controls the contribution of the offset sampling points, effectively minimizing the influence of irrelevant features.

\textit{3) Modulator:}
The deform block concentrates on extracting pixel-wise features of the LR and Ref image.
The global Mamba block emphasizes the extraction of patch-wise global features from the entire LR image, while the local Mamba block pays attention to the patch-wise local features from different parts of the Ref image.
Before the integration of deformable and Mamba features, we seek to further refine the ability of the model to perceive fine-grained details.
To realize this mechanism, we propose a modulator to refine and amplify the deformable features within the global LR feature of the global Mamba block, and the local Ref feature of the local Mamba block, respectively. Fig.~\ref{fig:Mamba}(C) depicts the structure of our modulator. The modulated Mamba features are selectively enhanced based on the output of a sigmoid function applied to their deformed counterparts. Due to the selective enhancement of Mamba features, the feature representation can be well-complemented by employing the modulator. Finally, modulated Mamba features and deformable features are combined to get the fused features.
\iffalse
The specific details are as follows:
\begin{equation}
F_{\text {GMamba}}^{lr\uparrow\prime}=\sigma\left(F_{\text {deform}}^{lr\uparrow}\right) \otimes F_{\text {GMamba}}^{lr\uparrow}
\end{equation}
\begin{equation}
F_{\text {LMamba}}^{ref\prime}=\sigma\left(F_{\text {deform}}^{ref}\right) \otimes F_{\text {LMamba}}^{ref}
\end{equation}
where $F_{\text {GMamba}}^{lr\uparrow\prime}$ is the modulated global Mamba features.
$F_{\text {LMamba}}^{ref\prime}$ is the modulated local Mamba features.
$\sigma(\cdot)$ denotes the sigmoid operation.
$\otimes$ represents the element-wise multiplication operation.
\fi
\begin{figure}[t]
    \centering
    \includegraphics[width=1\linewidth]{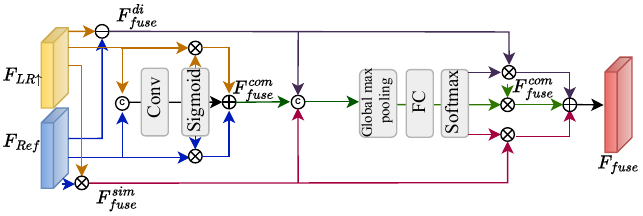}
    \caption{Architecture of our multi-modality feature fusion block.}
    \label{fig:Fusion}
\end{figure}

\subsection{Multi-Modality Feature Fusion Block}
The architecture of the multi-modality feature fusion block is depicted in the Fig.~\ref{fig:Fusion}.
Specifically, we input the LR features $F_{LR\uparrow}$ and Ref features $F_{Ref}$ to the designed multi-modality feature fusion block to perform three different types of fusion.
The fusion of different aspects of modalities focuses on capturing the differences between the $F_{LR\uparrow}$ and $F_{Ref}$ features by the element-wise subtraction operation, particularly those features that are prominently represented in one modality while being less pronounced or absent in the other.
This can suppress redundant features and further highlight critical differential features, as expressed below:
\begin{equation}
F_{fuse}^{di}= F_{LR\uparrow} - F_{Ref}. 
\end{equation}

For the fusion of similarity aspects of modalities, we use element-wise multiplication to emphasize the similarity between the $F_{LR\uparrow}$ and $F_{Ref}$ features. This enhances the capability of the module to identify consistent features across modalities, thereby facilitating a more effective fusion of high-quality reference image information. It is expressed as follows.
\begin{equation}
F_{fuse}^{sim}= F_{LR\uparrow} \otimes F_{Ref}. 
\end{equation}

We also explore the complementary information between modalities, which helps to mitigate the informational limitations of a single modality. We first use the convolution layer to extract features, and then apply the softmax function to normalize these features. The resulting weight map is divided into two independent weight maps, which are used to adjust the relative importance of the $F_{LR\uparrow}$ and $F_{Ref}$ features, respectively. This can be formulated as follows:
\begin{equation}
w_{LR\uparrow}, w_{Ref}=softmax\left(\operatorname{conv}\left(c a t\left(F_{LR\uparrow}, F_{Ref}\right)\right)\right), 
\end{equation}
\begin{equation}
F_{fuse}^{com}=\left(F_{LR\uparrow} \otimes w_{LR\uparrow}\right) \oplus\left(F_{Ref} \otimes w_{Ref}\right).
\end{equation}

As the image transitions from a severely degraded low-resolution to super-resolved image, the information gradually becomes more enriched at different stages. Consequently, the contributions of different fusion manners are dynamic.
Therefore, after obtaining the features of intermodal differences, similarity, and complementarity, we further fuse these three features using a weighted fusion method.
Initially, the three features are concatenated and passed through a global max pooling operation to extract the most representative feature value from each channel. The resulting pooled features are then fed into a fully connected layer to obtain normalized weights. The weights calculated by softmax are then used to perform a weighted sum on the original features, resulting in the final fused feature $F_{fuse}$.
The specific representation is as follows:
\begin{equation}
 w_{di}, w_{sim}, w_{com}=\operatorname{softmax}\left(f c\left(g m p\left(c a t\left(F_{fuse}^{di}, F_{fuse}^{sim}, F_{fuse}^{com}\right)\right)\right)\right),
\end{equation}
 \begin{equation}
 \quad F_{fuse}=\left(F_{fuse}^{di} \otimes w_{di}\right) \oplus\left(F_{fuse}^{sim} \otimes w_{sim}\right) \oplus\left(F_{fuse}^{com} \otimes w_{com}\right).
\end{equation}

\subsection{Objective Function}
 
We use the $\mathcal{L}_1$ loss function to quantify the difference between the super-resolved $SR$ and the ground truth with high resolution noted $HR$, as well as between reconstructed reference image $Rec_{Ref}$ and ground truth $Ref$ images. The $\mathcal{L}_1$ loss, also known as Mean Absolute Error (MAE), calculates the average of the absolute differences between the corresponding pixel values of the super-resolved and ground truth images, as shown below. 
\begin{equation}
	\begin{aligned}
		\mathcal{L}_{1}^{SR}(SR, HR)=\left|SR-HR\right|,
	\end{aligned}
\end{equation}
\begin{equation}
	\begin{aligned}
		\mathcal{L}_{1}^{Ref}(Rec_{Ref}, Ref)=\left|Rec_{Ref}-Ref\right|.
	\end{aligned}
\end{equation}

While $\mathcal{L}_1$ loss is highly effective in penalizing pixel-level errors, it primarily focuses on the differences between discrete pixels and often overlooks the crucial context and spatial relationships within the image. Therefore, we design a contrastive edge loss (CELoss) based on the Laplacian operator to guide the generation of more detailed super-resolved MR images by enhancing edge and local contrast information:

\begin{equation}
\mathcal{L}_{CELoss}= \frac{1}{3}\sum_{i=1}^{3}\left(\mathbf{E}_i \odot SR-\mathbf{E}_i \odot HR\right)^{2},
\end{equation}
\[
E_1 = \begin{bmatrix}
0 & -1 & 0 \\
-1 & 4 & -1 \\
0 & -1 & 0
\end{bmatrix}, \quad
E_2 = \begin{bmatrix}
-1 & 0 & -1 \\
0 & 4 & 0 \\
-1 & 0 & 1
\end{bmatrix}, \quad
E_3 = \begin{bmatrix}
1 & 1 & 1 \\
1 & -8 & 1 \\
1 & 1 & 1
\end{bmatrix}.
\]
$\mathrm{E}_i$ denotes the $i$ th contrastive edge convolution kernel. $i$ belongs to the interval $[0,2]$.
Specifically, E1 enhances both horizontal and vertical edges by analyzing the differences between neighboring pixels, making the horizontal and vertical details in the image more prominent.
E2 focuses on detecting diagonal edges by applying diagonal differential calculations, making these oblique edges clearer. E3 improves local contrast by comparing the central pixel with all its surrounding neighbors, making the details within local areas more noticeable.
The three kernels are specifically designed to enhance image edges and local contrast, bringing out the finer details in liquid and moisture regions within medical images.

The final loss function ensures that the $\mathcal{L}_1$ loss contributes to pixel-level reconstruction, while the CELoss focuses on restoring critical visual features related to edges and contrast. It is represented as:
\begin{equation}
\begin{split}
Loss = \alpha\mathcal{L}_{1}^{SR} + \beta\mathcal{L}_{1}^{Ref} + \gamma\mathcal{L}_{CELoss}, 
\end{split}
\label{equation15}
\end{equation}
where $\alpha,\beta$, and $\gamma$ are the balanced weight parameters.

\section{Experiments}\label{Experiments}
\subsection{Experimental Settings}

\textit{1) Datasets:}
We conduct our experiments using two medical image datasets: BraTS2021\footnote{http://braintumorsegmentation.org/} and IXI\footnote{https://brain-development.org/ixi-dataset/}.
BraTS2021 is the brain MRI dataset for multimodal brain tumor segmentation.
Following~\cite{DBLP:conf/iccv/LeiFZZ23}, the T1 modality is used as the reference image to guide the T2 modality for super-resolution. The slice resolution of BraTS2021 is 240$\times$240. We use 419 subjects for training and 156 for testing.
IXI dataset contains brain multi-modality MR image data. The proton density weighted imaging (PD) modality serves as the reference image to guide the T2 modality. The width and height of brain data are 256$\times$256. We use 368 subjects for training and 92 subjects for testing. 

\iffalse
\textbf{Hecktor} is designed for a challenge focused on segmentation and outcome prediction, specifically utilizing PET/CT images of the head and neck region.
The CT modality is utilized as the reference image to guide the super-resolution of the PET modality. In the BraTS2021 dataset, each slice has a resolution of 128$\times$128 pixels. 366 subjects are included in the training set, while 158 subjects are allocated for testing.
\fi

\textit{2) Evaluation Metrics:}
We assess the quality of the super-resolved images using the Peak Signal-to-Noise Ratio (PSNR) and the Structural Similarity Index (SSIM).
Further, we also use dice coefficient to validate the effectiveness of the downstream tumor segmentation task.

\textit{3) Experimental Details:}
Our approach is implemented with the PyTorch toolbox and trained on NVIDIA Tesla V100 GPU. The training low-resolution samples are generated by the frequency domain method~\cite{36}, ensuring a closer alignment with the actual distribution of low-resolution images. We adopt Adam as the optimizer with the initial learning rate of $2 \times 10^{-4}$ and batch size of 2. The numbers of deform block and Mamba block are set to 4. 
We partition the feature maps into non-overlapping 2$\times$2 patches using a sliding window with stride 2. This choice is supported by prior empirical results~\cite{DBLP:journals/prl/JiZKLVR25}. Using smaller patches of 2$\times$2 helps preserve fine details and supplies higher-resolution inputs to the Mamba block, thereby enhancing performance. In comparison, larger patches lose detail and impair performance.
The channel count is 96. Balancing parameters in Eq.\ref{equation15} are experimentally set to 0.7, 0.3, and 0.1, respectively.

\subsection{Ablation Study}

1) Effectiveness of the global Mamba for the low-resolution input:
To verify the effectiveness of the global Mamba for the single-modality low-resolution input, we use the deform block combined with local and global Mamba to process the low-resolution input, which are represented as \textit{LRLocalMamba} and \textit{LRGlobalMamba}, respectively.
The quantitative results are displayed in the first two rows of Tab.\ref{tab1}.
It can be seen that the \textit{LRGlobalMamba} has a higher performance than the \textit{LRLocalMamba}. The reason is that low-resolution images lose a significant amount of detail, especially in terms of textures, edges, and complex structures. Global Mamba can extract valuable information from the entire image, even in areas where detail is sparse. By utilizing the global context, they are better equipped to infer and restore these missing details. Conversely, local Mamba, which focuses predominantly on localized regions of the image, may struggle to accurately infer details when the information within these regions is insufficient, due to the lack of broader contextual cues.

\begin{table}[t]
\centering
\caption{Ablation study with different components in our approach.}
\setlength{\tabcolsep}{7mm}{
\begin{tabular}{c|cc}
\hline
Method                        & PSNR$\uparrow$  & SSIM$\uparrow$      \\ \hline
LRLocalMamba                  & 41.75 & 0.9899  \\
LRGlobalMamba                 & 41.91 & 0.9902  \\
Modulator       & 42.06 & 0.9904  \\
LRGlobal-RefGlobal                     & 42.25 & 0.9908  \\
LRGlobal-RefLocal             & 42.35 & 0.9909  \\
Multi-modality Feature Fusion & 42.43 & 0.9910  \\
Ours                          & 42.58 & 0.9912  \\ \hline
\end{tabular}}
\label{tab1}
\end{table}

2) Effectiveness of the modulator block:
To verify the effectiveness of the modulator block, we add it to the \textit{LRGlobalMamba}, denoted as \textit{Modulator}.
The third row of Tab.\ref{tab1} demonstrates that the modulator block enhances super-resolution performance. This verifies that the modulator block can refine and enhance the deformable features within the global feature of the global Mamba block.

3) Effectiveness of the local Mamba for reference input:
We also run the ablation experiment to verify the effectiveness of the local Mamba for reference input. 
The comparative results are presented in the fourth and fifth rows of Tab.\ref{tab1}. Compared with the \textit{Modulator}, we add the high-resolution reference image as a guide. We use the global and local Mamba to extract reference image features, which are expressed as \textit{LRGlobal-RefGlobal} and \textit{LRGlobal-RefLocal}, respectively.
From the PSNR and SSIM value comparison results, we can find that the module using local Mamba outperforms the module using global Mamba.
The explanation is that the global Mamba is better suited for capturing global contextual information, which may lead to excessive smoothing of the reference image features when dealing with reference images that already have high quality and sufficient contextual information. The local Mamba can retain and utilize the high-frequency details present in the reference image.

4) Effectiveness of the proposed multi-modality feature fusion block: To verify the effectiveness of the multi-modality feature fusion module, we replace the addition and 3$\times$3 convolution operation in \textit{LRGlobal-RefLocal} with multi-modality feature fusion block.
In the sixth line of Tab.\ref{tab1}, we report the quantitative comparisons with the \textit{Multi-modality Feature Fusion}. We also conduct additional ablation experiments for parameters and FLOPs. \textit{LRGlobal-RefLocal} has approximately 1.269M parameters and 37.05G FLOPS. In contrast, \textit{Multi-modality Feature Fusion} is a lighter model with 1.187M parameters and 32.27G FLOPS.

The proposed multi-modality feature fusion module achieves superior performance while introducing lower parameters and FLOPS compared to conventional convolution-based fusion methods. 
The original fusion module increases FLOPs and parameters due to its heavy 3$\times$3 convolution, despite using simple element-wise addition. In contrast, our designed fusion module captures cross-modal interactions efficiently via subtraction, multiplication, and lightweight spatial attention with multi-branch fusion and adaptive weighting. Using only a 1$\times$1 convolution for attention weights and mainly parameter-free operations, it entirely avoids 3$\times$3 convolutions and thereby achieves a much lower parameter count. Thanks to its powerful modeling ability, the effective information contained in the reference image can be fully explored.
\begin{figure}[t]
    \centering
    \includegraphics[width=1\linewidth]{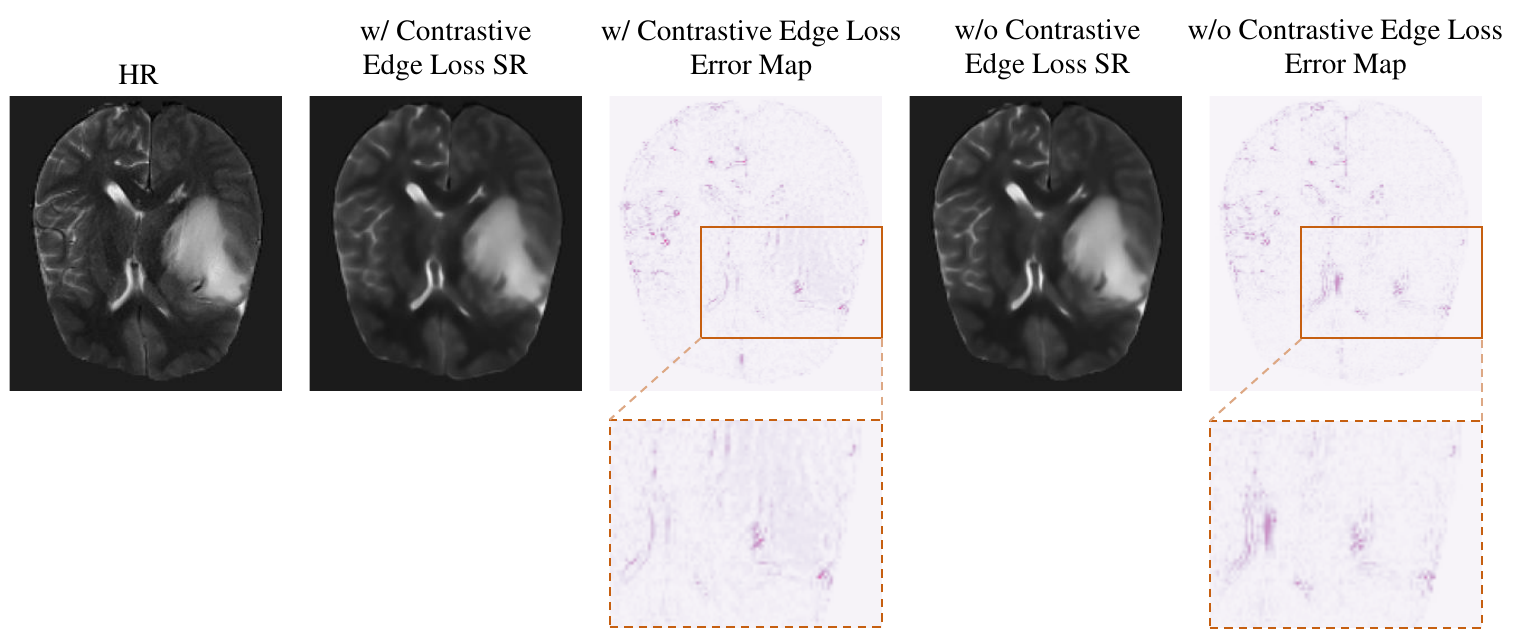}
    \caption{Qualitative results of super-resolved medical images and error map with and without contrastive edge loss on the BraTS2021 dataset under 4$\times$ unsampling factor.}
    \label{fig:EdgeAblation}
\end{figure}

5) Effectiveness of the contrastive edge loss:
Based on our \textit{Multi-modality Feature Fusion}, we conduct additional experiment to verify the
effectiveness of our designed contrastive edge loss function. The last row of Tab.\ref{tab1} tabulates the quantitative result. It can be found that the values of the two metrics get the best results than other modules. 
As shown in the Fig.\ref{fig:EdgeAblation}, models trained without the designed contrastive edge loss tend to produce smoother but blurrier edges. In contrast, models trained with the designed contrastive edge loss preserve sharper and more distinct structural details, especially around anatomical boundaries. The quantitative and qualitative results confirm the effectiveness of the contrastive edge loss function in enhancing edge fidelity during super-resolution.

\subsection{Comparison With Current Methods}
\textbf{1) Quantitative Analysis.}

\begin{table}[t]
\centering
\caption{Quantitative results with different methods on BraTS2021 dataset under 2$\times$ and 4$\times$ upsampling factor.}\label{tabbrats2021}
\setlength{\tabcolsep}{5mm}{
\begin{tabular}{c|cc|cc}
\hline
\multirow{2}{*}{Method} & \multicolumn{2}{c|}{BraTS2021 2$\times$} & \multicolumn{2}{c}{BraTS2021 4$\times$} \\  
                        & PSNR$\uparrow$           & SSIM$\uparrow$           & PSNR$\uparrow$          & SSIM$\uparrow$        \\ \hline
SRCNN                   & 37.88        &0.9794       & 30.44        & 0.9130     \\
T$^{2}$Net                   & 39.72      & 0.9850         & 32.48      & 0.9262     \\
DiVANet                     & {41.05}      & {0.9886}        & {34.31}      & {0.9498}      \\
MSHTNet                     & {39.45}      & {0.9841}        & {33.83}      & {0.9442}      \\
McMRSR  & {40.28}      & {0.9884}        & {34.72}      & {0.9557}      \\
MINet                     & {41.07}      & {0.9887}        & {34.83}      & {0.9563}      \\
Wavtrans                     & {42.02}      & {0.9904}        & {35.74}      & {0.9655}      \\
SANet                     & {41.39}      & {0.9893}        & {34.94}      & {0.9577}      \\
Ours                     & \textbf{42.58}       & \textbf{0.9912}        & \textbf{36.75}       & \textbf{0.9696}      \\ \hline
\end{tabular}}
\end{table}

\begin{table}[t]
\centering
\caption{Quantitative results with different methods on IXI dataset under 2$\times$ and 4$\times$ upsampling factor.}\label{tabIXI}
\setlength{\tabcolsep}{5mm}{
\begin{tabular}{c|cc|cc}
\hline
\multirow{2}{*}{Method} & \multicolumn{2}{c|}{IXI 2$\times$} & \multicolumn{2}{c}{IXI 4$\times$} \\  
                        & PSNR$\uparrow$           & SSIM$\uparrow$           & PSNR$\uparrow$          & SSIM$\uparrow$        \\ \hline
SRCNN                   & 29.23        & 0.8649       & 28.12        & 0.8357     \\
T$^{2}$Net                   & 31.31      & 0.9035         & 29.73      & 0.8773     \\
DiVANet                     & {33.15}      & {0.9320}        & {30.46}      & {0.8946}      \\
MSHTNet                     & {32.77}      & {0.9276}        & {30.44}      & {0.8963}      \\
McMRSR  & {36.08}      & {0.9624}        & {33.51}      & {0.9511}      \\
MINet                     & {36.22}      & {0.9618}        & {35.99}      & {0.9607}      \\
Wavtrans                     & {38.77}      & {0.9740}        & {37.99}      & {0.9708}      \\
SANet                     & {37.20}      & {0.9683}        & {36.01}      & {0.9604}      \\
Ours                     & \textbf{39.47}       & \textbf{0.9763}        & \textbf{38.23}       & \textbf{0.9717}      \\ \hline
\end{tabular}}
\end{table}

\begin{table}[t]
\centering
\caption{Param and FLOPS results with different multi-modality super-resolution methods.}\label{tabparam}
\setlength{\tabcolsep}{5mm}{
\begin{tabular}{c|clll}
\hline
Method & MINet  & Wavtrans & SANet & Our   \\ \hline
Param [M]  & 5.828  & 2.04     & 7.187 & 1.187 \\
FLOPS [G] & 327.05 & 63.89    & 404.7 & 32.27 \\ \hline
\end{tabular}}
\end{table}

\begin{figure} [t]
    \centering
    \includegraphics[width=1\linewidth]{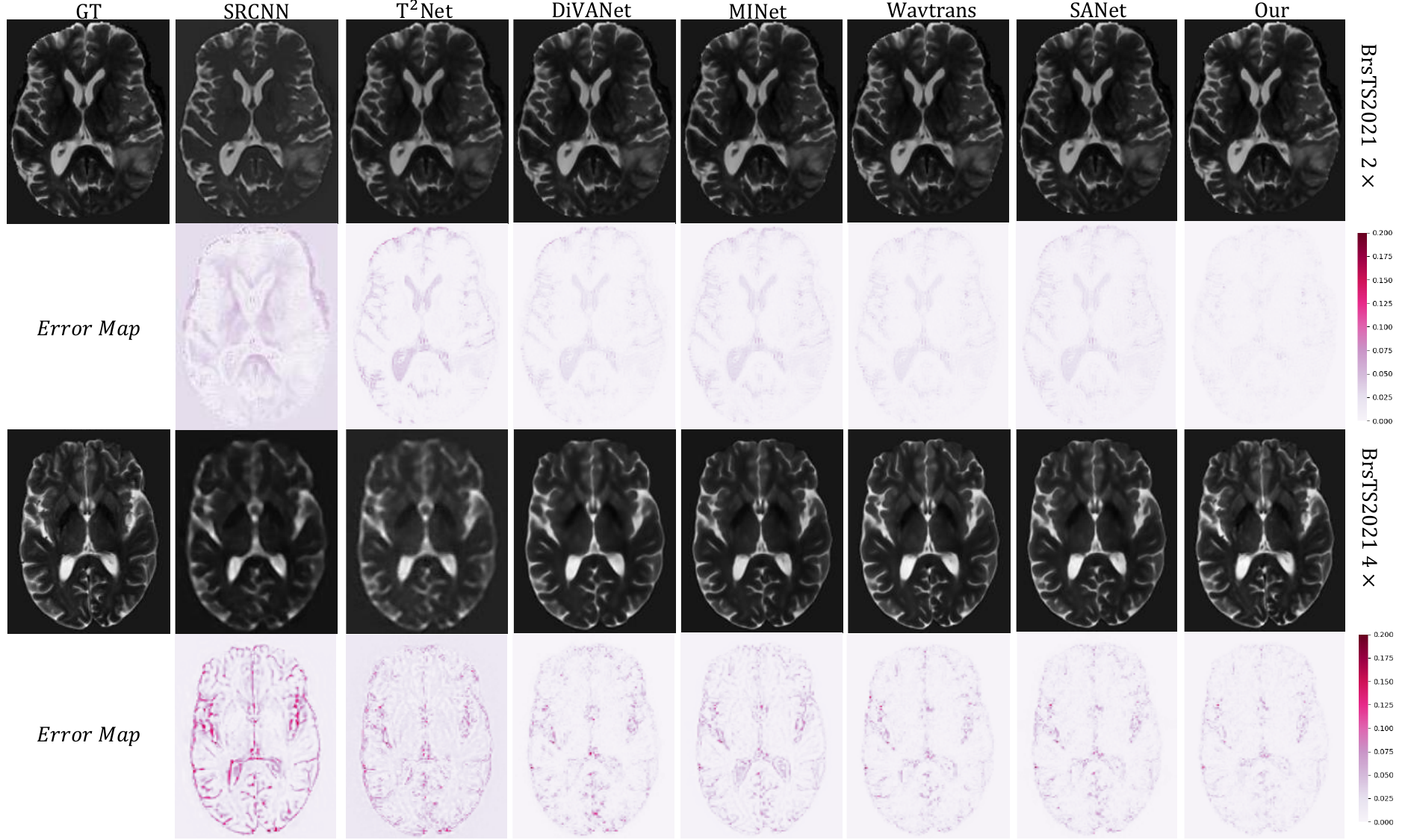}
    \caption{Qualitative results of super-resolved medical images on the BraTS2021 dataset under 2$\times$ and 4$\times$ unsampling factor.}
    \label{fig:BraTS}
\end{figure}

We compare our approach with some single-modality super-resolution methods (including SRCNN~\cite{8}, T$^2$Net~\cite{12}, DiVANet~\cite{DBLP:journals/pr/BehjatiRFHMG23}, and MSHTNet~\cite{liu2025remote}), and multi-modality super-resolution methods (including McMRSR~\cite{DBLP:conf/cvpr/LiLTDWX022}, MINet~\cite{DBLP:conf/miccai/FengFYX21}, WavTrans~\cite{DBLP:conf/miccai/LiLWDQ22}, and SANet~\cite{feng2024exploring}) on BraTS2021 and IXI dataset with 2$\times$ and 4$\times$ upsampling factors. 
Quantitative results are shown in Tab.\ref{tabbrats2021} and Tab.\ref{tabIXI}. We can see from the tables that 
most multi-modality super-resolution results outperform single-modality ones, demonstrating that high-resolution reference modalities can effectively guide the reconstruction of low-resolution images. Our approach achieves larger PSNR and SSIM values than most compared methods with 2$\times$ and 4$\times$ scale factors. In addition, it is noteworthy that our approach also has superior parameter efficiency compared to other multi-modality super-resolution methods from Tab.\ref{tabparam}.
The main reasons may lie in the following: 
(i) the designed global and local Mamba network can let the network efficiently learn patch-level local and global contextual information;
(ii) the multi-modality feature fusion module can effectively enhance feature learning between modalities for super-resolution;
(iii) the designed contrastive edge loss can further learn the edge and contrast-related features.

\begin{figure}[t]
    \centering
    \includegraphics[width=1\linewidth]{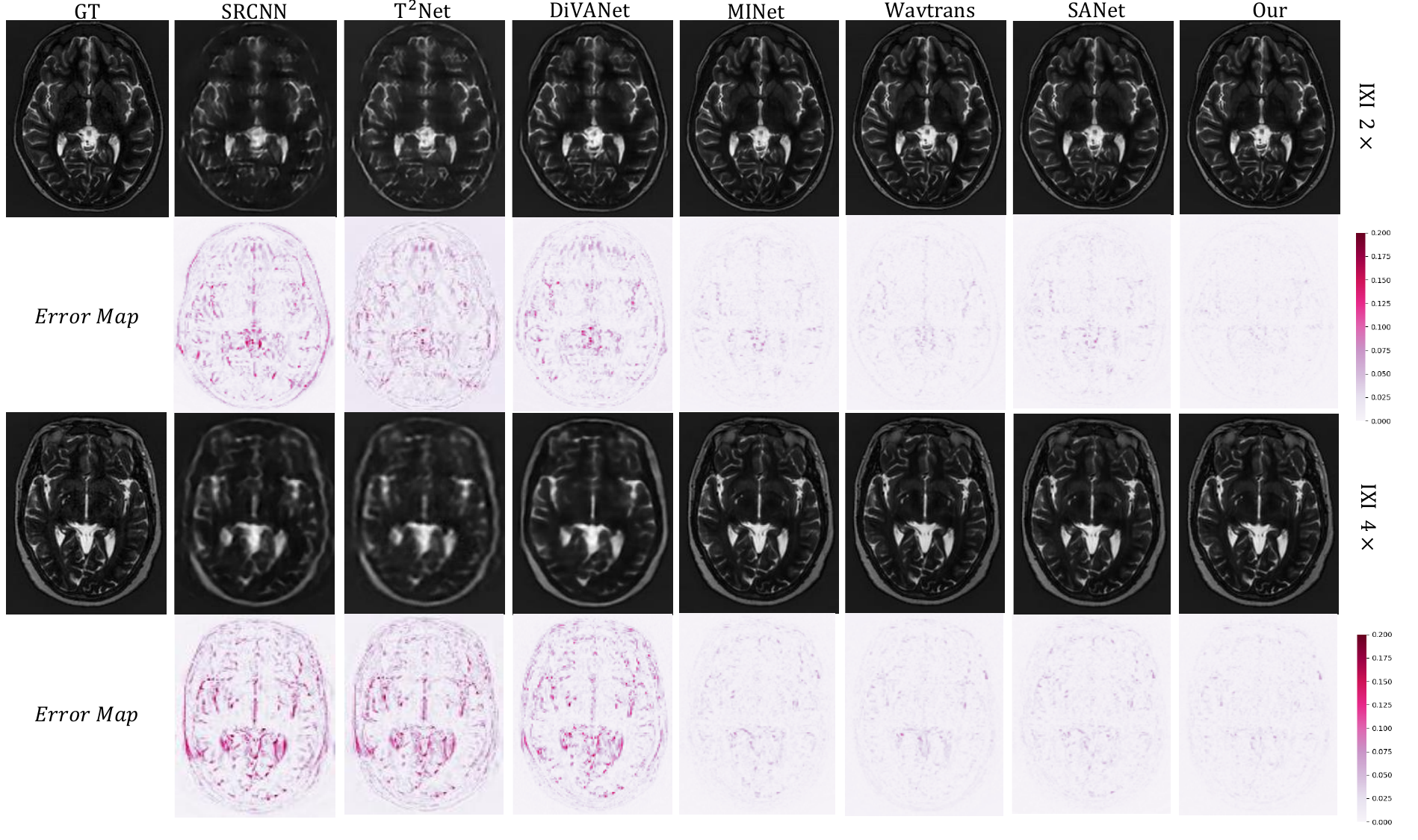}
    \caption{Qualitative results of super-resolved medical images on the IXI dataset under 2$\times$ and 4$\times$ unsampling factor.}
    \label{fig:IXI}
\end{figure}

\textbf{2) Qualitative Analysis.}
We also conduct qualitative comparisons to further validate our approach.
Fig.\ref{fig:BraTS} and Fig.\ref{fig:IXI} show the qualitative results of 2$\times$ and 4$\times$ upsampling factor on the BraTS2021 and IXI datasets, respectively. The first row shows the super-resolved images. To more clearly observe the differences with the ground truth images, we also present the corresponding error maps in the second row. The brighter the error map, the better the resolution of the super-resolved image, and vice versa.
As shown, the early proposed SRCNN fail to generate clear soft tissue structures. Although more recent work of the MINet, WavTrans, and SANet can recover the main structures, they fail to restore sharp textures and details. On the contrary, our approach can effectively recover more high-frequency details. 

\begin{table}[t]
\centering
\caption{Quantitative results of tumor segmentation.}\label{segmentation}
\setlength{\tabcolsep}{7mm}{
\begin{tabular}{c|c}
\hline
Method   & Dice Score (\%, ↑)   \\ \hline
MINet    & 66.159 \\
Wavtrans & 67.104 \\
SANet    & 66.322 \\
Our      & 67.219 \\
Initial Image       & 70.200 \\ \hline
\end{tabular}}
\end{table}

\begin{figure}[t]
    \centering
    \includegraphics[width=1\linewidth]{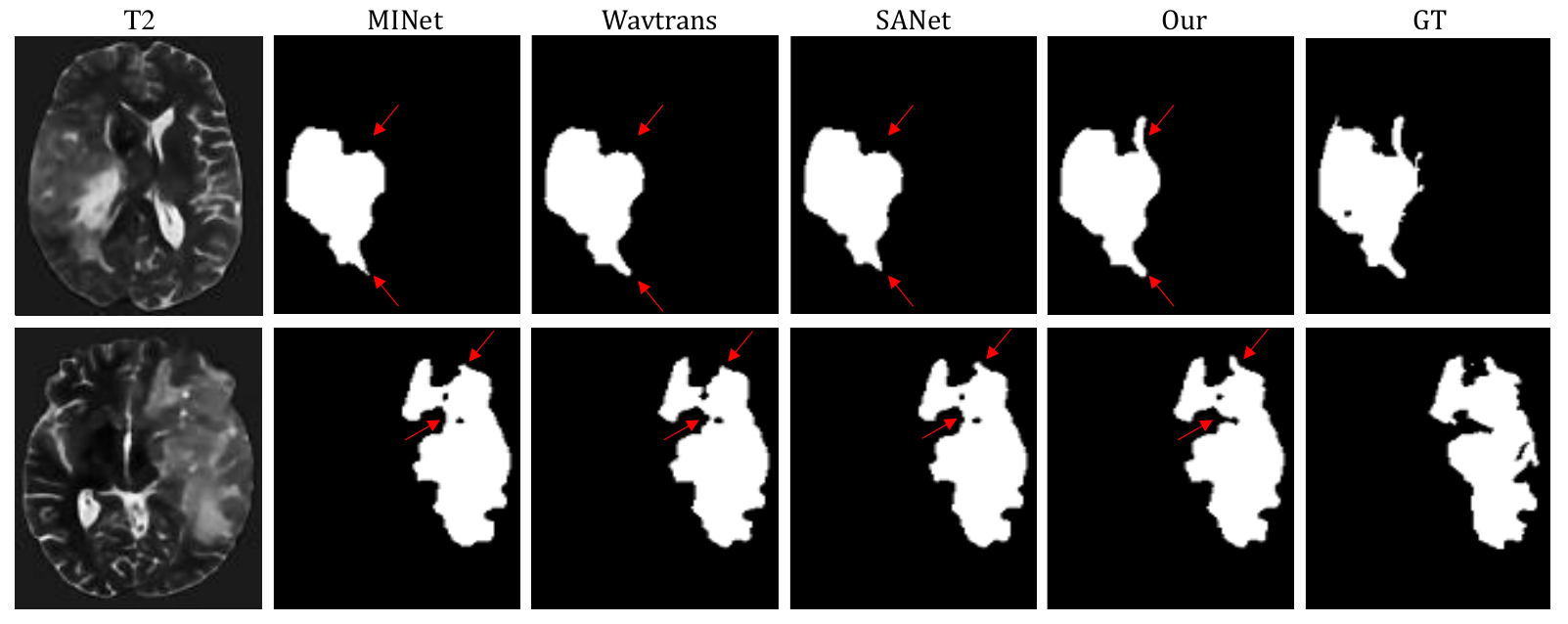}
    \caption{Qualitative results of tumor segmentation.}
    \label{fig:SegBraTS}
\end{figure}

\subsection{Effect on Downstream Segmentation Task}
High-resolution medical images can contain more details, which allows segmentation to identify the tumor region more precisely. The performance of downstream segmentation task is closely associated with the effect of super-resolution. Therefore, based on the tumor segmentation model of SwinUnet~\cite{DBLP:conf/eccv/CaoWCJZTW22}, we indirectly assess the performance of medical image super-resolution methods by evaluating the whole tumor segmentation on BraTS2021 dataset. 
Specifically, we compare our approach with three multi-modality super-resolution approaches: MINet, Wavtrans, and SANet.
As can be seen from Tab.\ref{segmentation} that our approach achieves the highest dice values, demonstrating the superiority of our method. The qualitative results are shown in Fig.\ref{fig:SegBraTS}.
The segmentation results of other methods are relatively rougher, and there are visible deviations from the ground truth, especially around the regions highlighted by the red arrows.
The super-resolved images generated by our approach are superior in tumor segmentation accuracy than other approaches, especially near the edges where the shape of the tumor are better captured.

\section{Conclusion and Future Work}\label{Conclusion}
In this paper, we developed a global and local Mamba network (GLMamba) for multi-modality medical image super-resolution.
Existing multi-modal approaches often fail to differentiate between the distinct roles of low-resolution images and high-resolution references in the super-resolution process. 
In contrast to these methods, our approach systematically exploits the complementary characteristics of both image types to maximize their respective contributions.
The global Mamba branch is responsible for capturing long-range dependencies between image patches in low-resolution inputs. The local Mamba branch focuses on analyzing short-range local patterns from high-resolution references. 
Furthermore, a deform block is utilized to adaptively extract features from pixel regions, allowing both branches to capture more flexible and dynamic information.
The modulator block is designed to further refine and enhance the deformable features within the global representations produced by the global Mamba block and the local representations extracted by the local Mamba block. Considering that the high-resolution reference image of another modality can provide effective information for low-resolution image super-resolution, we propose to enhance the feature learning ability by introducing a multi-modality feature fusion block, consisting of similarities, differences, and complementary information fusion between modalities. Thanks to the fusion block, the network can acquire more comprehensive features across modalities.
Furthermore, we propose a contrastive edge loss to enhance edge textures and contrast generation in medical images.
Comprehensive experiments on the BraTS2021 and IXI datasets show that our method outperforms others in super-resolution tasks. To further validate its effectiveness, we conducted additional experiments on brain tumor segmentation using the BraTS2021 dataset, confirming that the super-resolved images generated by our method improve segmentation accuracy.

However, this paper also has some limitations that suggest potential directions for future research.
First, our approach processes 2D medical images. By effectively leveraging the full 3D spatial information, it is possible to significantly improve the performance of medical image super-resolution. Second, accurate image fusion in multi-modality medical image super-resolution task relies on the proper alignment of the two modalities beforehand. 
While public datasets typically provide pre-registered images, clinical data often lacks such preprocessing. Therefore, developing a multi-task learning framework that integrates alignment with super-resolution represents a promising direction for future research.

\noindent\textbf{Acknowledgements.}
The work was supported by the National Key R$\&$D Program of China (No.2018AAA0102100); the National Natural Science Foundation of China (Nos.U22A2034, 62177047); the Key Research and Development Program of Hunan Province (No.2022SK2054); Major Program from Xiangjiang Laboratory under Grant 23XJ02005; Central South University Research Programme of Advanced Interdisciplinary Studies (No.2023QYJC020); the Fundamental Research Funds for the Central Universities of Central South University (No.2024ZZTS0486); the China Scholarship Council (No.202306370195).

\bibliography{mybibfile}

\end{document}